\documentclass[sigconf]{acmart}

\usepackage{algorithm}
\usepackage{algorithmic}

\AtBeginDocument{%
  \providecommand\BibTeX{{%
    \normalfont B\kern-0.5em{\scshape i\kern-0.25em b}\kern-0.8em\TeX}}}

\copyrightyear{2021}
\acmYear{2021}
\setcopyright{acmcopyright}\acmConference[KDD '21]{Proceedings of the 27th ACM SIGKDD Conference on Knowledge Discovery and Data Mining}{August 14--18, 2021}{Virtual Event, Singapore}
\acmBooktitle{Proceedings of the 27th ACM SIGKDD Conference on Knowledge Discovery and Data Mining (KDD '21), August 14--18, 2021, Virtual Event, Singapore}
\acmPrice{15.00}
\acmDOI{10.1145/3447548.3467334}
\acmISBN{978-1-4503-8332-5/21/08}

\begin{document}

\title{ImGAGN:Imbalanced Network Embedding via \\ Generative Adversarial Graph Networks}


\author{Liang Qu}
\authornote{Both authors contributed equally to this research.}
\email{qul@mail.sustech.edu.cn}
\author{Huaisheng Zhu}
\authornotemark[1]
\affiliation{%
  \institution{Southern University of Science and Technology}
  \city{Shenzhen}
  \country{China}
  \postcode{518055}
}

\author{Ruiqi Zheng}
\affiliation{%
  \institution{Southern University of Science and Technology}
  \city{Shenzhen}
  \country{China}
  \postcode{518055}
}

\author{Yuhui Shi}
\authornote{Co-corresponding author.}
\email{shiyh@sustech.edu.cn}
\affiliation{%
  \institution{Southern University of Science and Technology}
  \city{Shenzhen}
  \country{China}
  \postcode{518055}
}

\author{Hongzhi Yin\textsuperscript{\dag}}
\email{h.yin1@uq.edu.au}
\affiliation{%
  \institution{The University of Queensland}
  \city{Brisbane, QLD 4072}
  \country{Australia}
}

\renewcommand{\shortauthors}{Qu and Zhu, et al.}

\begin{abstract}
Imbalanced classification on graphs is ubiquitous yet challenging in many real-world applications, such as fraudulent node detection. Recently, graph neural networks (GNNs) have shown promising performance on many network analysis tasks. However, most existing GNNs have almost exclusively focused on the balanced networks, and would get unappealing performance on the imbalanced networks. 
To bridge this gap, in this paper, we present a generative adversarial graph network model, called ImGAGN to address the imbalanced classification problem on graphs. It introduces a novel generator for graph structure data, named GraphGenerator, which can simulate both the minority class nodes' attribute distribution and network topological structure distribution by generating a set of synthetic minority nodes such that the number of nodes in different classes can be balanced. Then a graph convolutional network (GCN) discriminator is trained to discriminate between real nodes and fake (i.e., generated) nodes, and also between minority nodes and majority nodes on the synthetic balanced network. To validate the effectiveness of the proposed method, extensive experiments are conducted on four real-world imbalanced network datasets. Experimental results demonstrate that the proposed method ImGAGN outperforms state-of-the-art algorithms for semi-supervised imbalanced node classification task.
\end{abstract}




\maketitle

\section{Introduction}
Network data, consisting of nodes (objects) and edges (objects' relationships), is ubiquitous in many real-world problems, such as social networks, protein-protein interaction networks, citation networks and so on. Recently, network embedding \cite{cai_comprehensive_2017,wu_comprehensive_2019,chen2019exploiting} techniques, which map the nodes of the original networks into the dense and low-dimensional vectors (called node embeddings) and preserve the network structure information as much as possible, have shown promising performance on many network data analysis tasks, such as node classification \cite{kipf_semi-supervised_2016,sun2021heterogeneous}, link prediction \cite{grover_node2vec:_2016,10.1145/3366423.3380073}, community detection \cite{fortunato2010community} and so on.

Typical network embedding methods could be roughly divided into two categories, unsupervised network embedding methods \cite{9185532} and semi-supervised network embedding methods \cite{8790139}. The former obtains the node embeddings by preserving the network structure information. Representative method like DeepWalk \cite{perozzi_deepwalk:_2014} utilizes the truncated random walks strategy to preserve network local information. The latter, semi-supervised network embedding methods, utilizes not only network structure information but also nodes' label information. Representative method like GCN \cite{kipf_semi-supervised_2016} obtains the target node embeddings by aggregating the neighbor nodes' feature information.

\begin{figure}[t]
\centering
\includegraphics[width=0.45\textwidth]{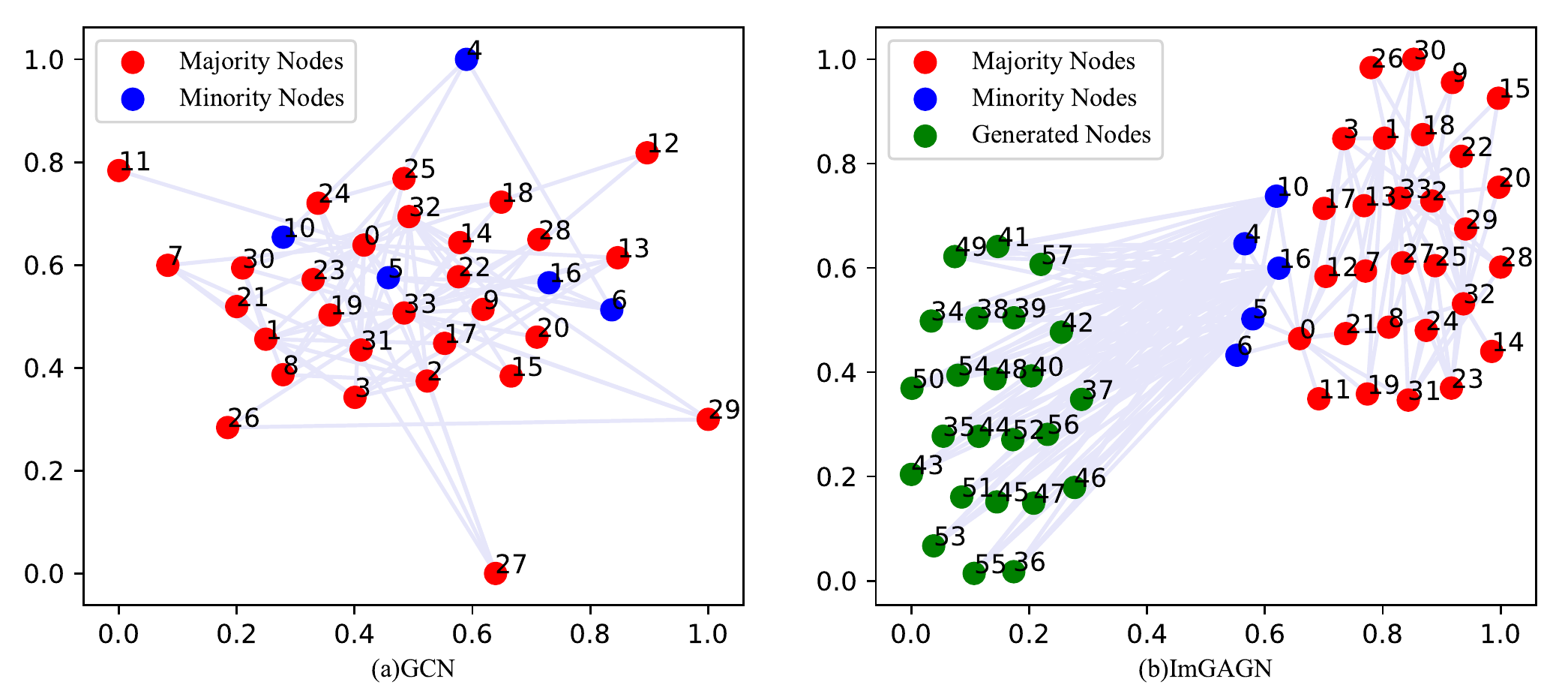} % Reduce the figure size so that it is slightly narrower than the column.
\caption{The 2-dimensional network embedding for the imbalanced network (Zachary's Karate Network \cite{zachary_information_1977}) using: (a) GCN \cite{kipf_semi-supervised_2016} (b) the proposed ImGAGN is capable of discriminating between the real nodes (i.e., the blue circles and red circles) and the generated fake nodes (i.e., the green circles), and also between the minority nodes (i.e., the blue nodes) and the majority nodes (i.e., the red nodes).}
\label{fig2}
\end{figure}

However, the extensive existing network embedding methods assume that the nodes' labels are balanced, i.e., every class has roughly equal number of examples. Generally, these methods could not obtain good performance on the imbalanced networks in which the number of examples of one class (minority) is far less than that of other classes (majority), and the minority usually plays an essential role in the real-world problems. For example, for the fraudulent node detection in the online social networks, the number of fraudsters is far less than that of the normal users, and the fraudsters often try to disguise their identities as the normal users. Therefore, two key challenges of imbalanced network analysis are that: (1) The number of one class examples (minority nodes) is far less than that of other classes (majority nodes) in the network, and the labeling for minority nodes is extremely expensive. (2) The minority nodes are non-separability from the majority nodes, that is, it is difficult to find the support regions of majority and minority nodes in the networks (as shown in Figure 1(a)).   

To address the above challenges,
% we first propose general framework for incorporating GANs into GNNs for semi-supervised imbalanced network embedding. Based on this framework,
we propose a semi-supervised generative adversarial graph network model, called ImGAGN. It introduces a GraphGenerator which can simulate both the minority class node's attribute distribution and network topological structure distribution by generating a set of minority class nodes linking to the real minority nodes to balance the original network classes distribution, then a GCN discriminator is trained to discriminate between real nodes and fake nodes, and also between minority nodes and majority nodes on the synthetic balanced network. Specifically, as shown in the Figure 1(b), the GraphGenerator iteratively learns to generate a set of minority nodes (green circles in Figure 1(b)) to make the original network classes balanced. The topological structure features of the generated nodes are obtained by linking the fake nodes to the real minority nodes (blue circles in Figure 1(b)) of the original network, and the attribute features of the generated nodes are obtained by averaging their neighbor nodes' (i.e., the real minority nodes) attribute features. Then the discriminator (GCN) is trained to discriminate whether the node is generated by generator and whether the node is minority class. From Figure 1, we can find that ImGAGN could generate a set of appropriate minority nodes to make the original minority nodes separate from the majority nodes, and the generated fake nodes separate from real nodes. We evaluate our proposed method on four publicly available real-world imbalanced datasets on semi-supervised imbalanced node classification task. Experimental results demonstrate that ImGAGN outperforms the state-of-the-art algorithms including both balanced network embedding methods and imbalanced network embedding methods. It is worth emphasizing that the GraphGenerator to generate the new balanced network is done after training/testing split, that is, the generated fake nodes would only be linked to the training minority nodes, but not the testing minority nodes. 

The main contributions of this paper are summarized as follows:

\begin{itemize}
    \item In this paper, we propose a novel semi-supervised generative adversarial graph network model, called ImGAGN, which utilizes a generator to simulate the minority class node distribution and generates a set of minority nodes to make original network classes balanced. Then GCN is trained to discriminate between the majority and minority nodes, and also between the fake nodes and real nodes on the synthetic balanced network classes.
    \item Based on ImGAGN, we propose a novel generator for graph structure data, called GraphGenerator, which can effectively learn not only the nodes' attribute feature distribution but also the network topological structure distribution.
    \item The proposed method is validated on four real-world imbalanced network datasets for imbalanced binary node classification and network layouts tasks. Experimental results demonstrate that the proposed method is superior to the state-of-the-art both balanced network embedding and imbalanced network embedding techniques. In addition, we released our codes to facilitate further researchers by others.\footnote{https://github.com/Leo-Q-316/ImGAGN.} 
\end{itemize}

The rest of the paper is organized as follows. Section 2 will introduce some main related works. Section 3 will formulate the problem and provide a detailed introduction to the proposed method. In Section 4, we will introduce the experimental setups and results followed by the conclusions in Section 5. 

\section{Related Works}
In this section, we introduce two main related research fields including imbalanced learning and imbalanced network embedding.
\subsection{Imbalanced learning}
Imbalanced learning techniques \cite{5128907,johnson2019survey} aim at solving the problem with imbalanced data in which at least the number of one class data (minority) is far less than that of other classes (majority). Generally speaking, the minority class is often high-impact on many real-world problems, such as the cancer detection in medical diagnosis and fraud detection in financial system. 

Existing methods for imbalanced learning mainly include: (1) sampling based methods, which learn the imbalanced classification by oversampling \cite{han2005borderline} the minority class or undersampling \cite{liu2008exploratory} the majority class. Representative method like SMOTE \cite{chawla_smote_2002} generates artificial data from existing minority class. (2) cost-sensitive learning based methods \cite{elkan2001foundations,ting2002instance}, which utilize different cost matrices for calculating the cost of any particular data examples misclassified. (3) kernel-based methods \cite{akbani2004applying}, which employ classifier like support vector machines (SVMs) \cite{suykens1999least} to maximize the separation margin. and (4) GANs based methods \cite{shamsolmoali_imbalanced_2020,montahaei_adversarial_2018,douzas2018effective}, which are similar to our proposed method using the generator to create the minority class for balancing the data classes distribution. However, to our best knowledge, little work has employed these GANs based methods to the imbalanced network data.

% \subsection{Network Embedding}
% Network embedding techniques aim at mapping the nodes of the original networks into the low-dimensional dense vector space and preserving the network structure information as much as possible. In this section, we briefly review several representative network embedding methods including both balanced and imbalanced setting.
% \subsubsection{Balanced network embedding}

\subsection{Imbalanced network embedding}
Imbalanced network embedding methods aim at solving the imbalanced learning problems on graph structure data.
GRADE \cite{he_graph-based_nodate} is the classic method for imbalanced network embedding. It utilizes the global similarity matrix to obtain the compact minority class clusters, and learns the decision boundary between majority and minority classes by selecting the examples from the regions where the density changes the most. Wu et al. \cite{wu_imverde:_2018} propose a novel random walk strategy, called vertex-diminished random walk (VDRW), which discourages the random particle to the nodes visited. Based on VDRW, they introduce the semi-supervised network embedding method ImVerde which consists of the context sampling and the balanced-batch sampling strategies to improve the quality of the node-context pairs. SPARC \cite{zhou_sparc:_2018} obtains the imbalanced node embedding in a mutually way, which can jointly predict the minority class and the neighbor context in the networks. RSDNE \cite{wang_rsdne_nodate} explores the network embedding with completely-imbalanced labels. It learns the imbalanced node embedding by allowing the intra-class nodes on the same manifold in the embedding space and removing the known connections between the inter-class nodes. DR-GCN \cite{shi_multi-class_2020} proposes two types of regularization to tackle imbalanced network embedding. It utilizes a conditional adversarial training to discriminate the nodes from different classes, and a distribution alignment training is employed to balance the majority and minority classes learning.

\begin{figure*}[]
    \centerline{\includegraphics[width=0.8\textwidth]{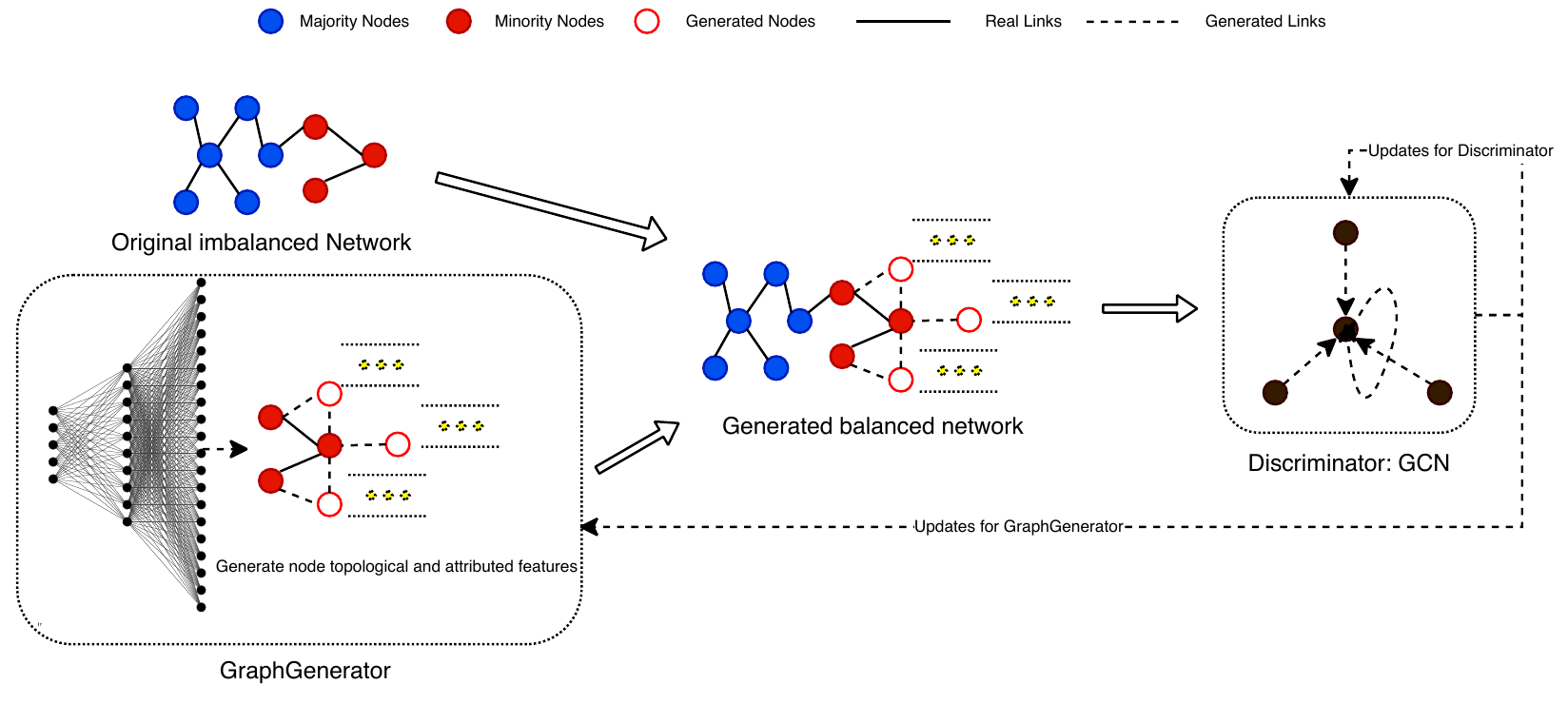}}
    \caption{The architecture of ImGAGN. The minority and majority nodes of original imbalanced network are represented by red and blue solid circles respectively, and the synthetic minority nodes generated by GraphGenerator are represented by red hollow circles in artificial synthetic classes balanced network. In addition, The links between real nodes are represented by solid lines, and the links between synthetic minority nodes and real minority nodes are represented by dashed lines.}
    \label{fig}
\end{figure*}

\section{Proposed Method}
In this section, we first provide several needed concepts related to the proposed method. Then, we present our proposed method ImGAGN in detail. Finally, we analyze the time complexity of the proposed method.
\subsection{Preliminary}
Before presenting our proposed ImGAGN, we provide a brief introduction to the needed concepts for proposing our method.
\begin{itemize}
    \item \textbf{Imbalanced network:} given an imbalanced network $\mathcal{G}_{im}=(V,E,A,X,C)$, where $V$ is the set of $n$ nodes, $E$ is the set of edges, $A$ is the adjacency matrix,  $X \in R^{n \times f}$ is the node feature matrix with feature dimension $f$, and $C = \{c_{min},c_{maj}\}$ is the set of node classes. $|c_{min}|$ and $|c_{maj}|$ represent the number of nodes in their classes. The network $\mathcal{G}_{im}=(V,E,A,X,C)$ is an imbalanced network if $|c_{min}|$ is far less than  $|c_{maj}|$ (i.e., $|c_{min}| \ll |c_{maj}|$).
    \item \textbf{Imbalanced network embedding:} imbalanced network embedding aims at mapping the node $v_{i} \in V$ of an imbalanced network $\mathcal{G}_{im}=(V,E,A,X,C)$ into a continuous low-dimensional vector $\vec{h_{i}} \in R^{d}$ ($d \ll n$), such that the nodes with the same class label are closer than the nodes with the different class labels in the embedding space.
    \item \textbf{GANs:} GANs \cite{goodfellow2014generative,gui2020review,wang2019enhancing,wang2018neural,yu2019generating} are a class of neural networks which consist of a generator and a discriminator. The key idea of generator $G$ is that it aims at generating the fake data to simulate the real data distribution to confuse discriminator. The goal of discriminator $D$ is to correctly classify both the real training data and fake data generated from generator $G$. The GANs methods can be formulated as follows \cite{goodfellow2014generative}:
    \begin{equation}\begin{array}{l}
    \min _{G} \max _{D} V(D, G)=E_{x \sim p_{\text {data }}(x)}[\log D(x)] \\
    \quad+E_{z \sim p_{z}(z)}[\log (1-D(G(z)))]
    \end{array}\end{equation}
    where $x$ is the real data obeying the distribution $p_{data}$, and $z$ is the noise variable obeying the distribution $p_{z}$.
\end{itemize}

% \subsection{A general imbalanced semi-supervised classification framework}
% Recent years, GANs  based imbalanced classification methods \cite{shamsolmoali_imbalanced_2020,montahaei_adversarial_2018,noauthor_effective_2018,douzas2018effective} have shown great advantages over the traditional oversampling methods (e.g., SMOTE \cite{chawla_smote_2002}) which generate synthetic data along the line segment of the minority class data. The reason is that, unlike traditional methods generating synthetic data from local information, GANs could learn the global minority class distribution to generate synthetic data. However, to our best knowledge, little work utilizes the GANs based methods to solve the imbalanced network classification problem. To bridge this gap, we first present a general framework for imbalanced learning with GAN. The key ideas of this framework are that a generator is utilized as a novel oversampling method generating synthetic data from the overall minority class distribution for balancing the original imbalanced data. Then a task-related discriminator is trained to classify both real data and generated data.
\subsection{ImGAGN}
To address the imbalanced classification problems on graph, we propose a GANs based imbalanced learning method, called ImGAGN, which incorporates GCN with a novel generator named GraphGenerator for graph structure data. It generates a set of synthetic minority nodes such that the number of nodes in different classes can be balanced. In addition, GraphGenerator can effectively learn not only the nodes' attribute distribution but also the network topological structure distribution. Then GCN discriminator is trained to discriminate between real nodes and fake (i.e., generated) nodes, and also between minority nodes and majority nodes on the synthetic balanced network. The architecture of ImGAGN is shown in Figure 2.

\subsubsection{\textbf{GraphGenerator (G)}}
Unlike traditional GAN processing regular Euclidean data (e.g., images and text) which data is dependent on each other, the generator only need to learn the data feature distribution. For graph structure data, because the data (i.e., nodes) is independent of each other, the generator needs to learn not only nodes' attribute features distribution (e.g., the node features) but also network topological structure distribution (e.g., the node link relationships). In this paper, we propose a novel generator for graph data, called GraphGenerator, which can generate the node link relationships between the synthetic minority nodes and the real minority nodes, and the features of the synthetic minority nodes are obtained by averaging the features of the linked real minority nodes. 

GraphGenerator $G_{graph}: \mathcal{Z} \to \mathcal{F} \times \mathcal{T} $ is a fully connected neural network, where $\mathcal{Z}$ is the noise space with $d_{z}$ dimension, and $\mathcal{F},\mathcal{T}$ are network feature space and network structure space respectively. Specifically, for an imbalanced network $\mathcal{G}_{im}=(V,E,A,X,C)$, let $n_{maj}$ and $n_{min}$ represent the number of majority nodes and the number of minority nodes respectively with $n = n_{maj}+n_{min}$. Let $n_{g} = n_{maj}-n_{min}$ represents the number of nodes needing to be generated for balancing the network classes distribution. Thus, the number of units in input layer is $d_{z}$, and the number of units in output layer is $d_{o} = n_{g} \times n_{min}$. For better understanding, we convert the output vector $\vec{o} \in R^{d_{o}}$ into the matrix form $O \in R^{n_{g} \times n_{min}}$, and then we apply $softmax(O_{i})$ function to normalize each row in $O$ as equation (2):

\begin{equation}
T_{i} = softmax(O_{i}) = \frac{e^{O_{ij}}}{\sum_{j=1}^{n_{min}}e^{O_{ij}}}, for i=1,...,n_{g}
\end{equation}

where each row $O_{i}$ represents the link relationship between each generated minority node to all real minority nodes. In addition, each element $T_{ij}$ represents the link normalized weight between the generate node $u_{i} \in U$ and original minority node $v_{j} \in V$
, where $U$ is the set of generated minority nodes. Thus, $T$ represents the networks topological structure information between generated minority nodes and original minority nodes.

% and each element $o_{ij} \in O$ is discretized into the $\{0,1\}$ space by function $Dis$ with the hyperparameter $n_{min}$ as equation (2):
% \begin{equation}
% b_{ij} = Dis(o_{ij})=\left\{\begin{array}{ll}
% 1, &  o_{ij} > \frac{1}{n_{min}} \\
% 0, &  o_{ij} \le \frac{1}{n_{min}}
% \end{array}\right., b_{ij} \in B
% \end{equation}
% where $B \in \{0,1\}^{n_{g} \times n_{min}}$ is the structure features of the generated minority nodes by GraphGenerator. Each row of $B$ represents the link relationships between each generated minority node to all real minority nodes, where $1$ represents link, and $0$ represents unlink. 

To generate the nodes' attribute features $X_{g} \in R^{n_{min} \times f}$ of the generated minority nodes, we aggregate the neighbor nodes' attribute features of each generated minority nodes as equation (3):

\begin{equation}
    X_{g} = TX_{min}
\end{equation}
where $X_{min} \in R^{n_{min} \times f} \subset X$ is the real minority nodes' features matrix of the original imbalanced network $\mathcal{G}_{im}$. And $f$ is the dimension of original minority node features.

\subsubsection{\textbf{Discriminator (D)}}
In this paper, we utilize the two-layer GCN \cite{kipf_semi-supervised_2016} as our discriminator, and the input of GCN is the new network $\mathcal{G}_{bal}=(V^{'},E^{'},A^{'},X^{'},C^{'})$ with balanced classes distribution by incorporating the generated minority nodes from GraphGenerator into the original imbalanced networks $\mathcal{G}_{im}$, where $V^{'}$ represents the new nodes set which consists of the nodes in $\mathcal{G}_{im}$ and the generated minority nodes by GraphGenerator, $E^{'}$ represents the new edges set which consists of the all edges in $\mathcal{G}_{im}$ and the generated edges by GraphGenerator, $A^{'}$ and $X^{'}$ are the new adjacency matrix and feature matrix associated to $V^{'}$ respectively.
$C^{'}=\{(real,minority),(real,majority),(fake,minority)\}$ represents the node labels set. It is worth noting that the GraphGenerator do not generate the majority nodes, thus the label $(fake, majority)$ is not included to $C^{'}$.

The goal of discriminator is to discriminate whether the nodes are generated by generator (i.e., fake) and also whether the node is minority class. Therefore, we can utilize the GCN as a node multi-class classification classifier, and the output $Y$ of GCN is calculated by equation (4) \cite{kipf_semi-supervised_2016} as follows:
\begin{equation}
    Y = softmax(\widehat{A^{'}}ReLU(\widehat{A^{'}}X^{'}\Omega^{0})\Omega^{1})
\end{equation}
where $\widehat{A^{'}}=\widehat{D^{-\frac{1}{2}}}(\widehat{A^{'}}+I_{N})\widehat{D^{-\frac{1}{2}}}$ is the pre-processing step following \cite{kipf_semi-supervised_2016} with identity matrix $I_{N}$ and $D_{ij}=\sum_{j}A_{ij}$. $\Omega^{0}$ and $\Omega^{1}$ are input-to-hidden and hidden-to-out weight matrices respectively.

\subsubsection{\textbf{Model Optimization}}

The loss function of the GraphGenerator is as equation (5). 
\begin{equation}\begin{array}{l}
    \mathcal{L}_{gen} = \mathcal{L}_{rf} + \mathcal{L}_{mi} + \mathcal{L}_{di} + \mathcal{L}_{re} \\
    \quad = \sum_{i=1}^{n_{g}}-q_{i}log Pr(\widehat{y_{i}}=real | \vec{x_{i}}) \\ 
    \quad + \sum_{i=1}^{n_{g}}-q_{i}log Pr(\widehat{y_{i}}=minority | \vec{x_{i}}) \\
    \quad + \frac{1}{|n_{g}|} \sum_{i=1}^{n_{g}}\sum_{j=1}^{n_{min}} ||\vec{x_{i}}-\vec{x_{j}}||_{2}^{2} \\
    \quad + \alpha||\Theta||_{2}^{2}
    \end{array}
\end{equation}
where this loss function consists of four terms. The first $\mathcal{L}_{rf}$ and second terms $\mathcal{L}_{mi}$ are the confusing discriminator loss over the generated minority data, in which $q_{i} \in C^{'}$ and $\widehat{y_{i}} \in Y$ denotes the ground-truth labels and the output (prediction probability) of the discriminator respectively, and $\vec{x_{i}}$ is the node embedding vector. The third term $\mathcal{L}_{di}$ aims at making the generated minority nodes close to the real minority nodes. The last term $\mathcal{L}_{re}$ is regularizer, in which $\Theta$ is the set of training weights of GraphGenerator with regularization coefficient $\alpha$.

The loss function of discriminator is as equation (6):
\begin{equation}\begin{array}{l}
    \mathcal{L}_{dis} = \mathcal{L}_{fa} + \mathcal{L}_{cl} + \mathcal{L}_{mm} + \mathcal{L}_{ree} \\
    \quad = \sum_{i=1}^{n_{g}+n_{min}+n_{maj}} -[q_{i}log(Pr(\widehat{y_{i}}=fake | \vec{x_{i}})) \\ +(1-q_{i})log(1-Pr(\widehat{y_{i}}=fake | \vec{x_{i}}))] \\
    \quad + \sum_{i=1}^{n_{g}+n_{min}+n_{maj}} -[q_{i}log(Pr(\widehat{y_{i}}=minority | \vec{x_{i}})) \\ +(1-q_{i})log(1-Pr(\widehat{y_{i}}=minority | \vec{x_{i}}))] \\
    \quad - \sum_{i=1}^{n_{min}}\sum_{j=1}^{n_{maj}}||\vec{h_{i}}-\vec{h_{j}}||_{2}^{2} \\
    \quad + \beta||\Omega||_{2}^{2}
    \end{array}
\end{equation}
where this loss function consists of four terms. The first term $\mathcal{L}_{fa}$ is the cross entropy loss to discriminate that the node is generated by generator or real node of the original network. The second term $\mathcal{L}_{cl}$ is also the cross entropy loss to discriminate that the node is minority class or majority class. The third term $\mathcal{L}_{mm}$ aims at making the embeddings of the different class nodes are far away from each other. The last term $\mathcal{L}_{ree}$ is regularizer, in which $\Omega$ is the set of training weights of the discriminator with regularization coefficient $\beta$.

Finally, the adversarial training objective function of ImGAGN is given as equation (7):
    \begin{equation}\begin{array}{l}
    \min _{G} \max _{D} V(D, G)=E_{x \sim p_{\text {data }}(x)}[\log D(x)+\mathcal{L}_{cl}+\mathcal{L}_{mm} + \mathcal{L}_{ree}] \\
    \quad+E_{z \sim p_{z}(z)}[\log (1-D(G(z)))+\mathcal{L}_{mi}+\mathcal{L}_{di} + \mathcal{L}_{re}]
    \end{array}\end{equation}
    
The goal of GraphGenerator is to generate the fake minority nodes to simulate the real minority nodes distribution to confuse discriminator. The goal of discriminator is to correctly classify between the real training nodes and the fake nodes generated from GraphGenerator, and also between the minority nodes and the majority nodes.

\subsection{Time Complexity}
The time complexity of the proposed ImGAGN is as follows. The complexity for updating generator is $O((L-1)n_{g}H^{2}+n_{g}n_{min}^{2})$, where $L$ is the number of fully connected layers of generator, 
% $n_{g}$ and $n_{min}$ are the generated    and real minority nodes number respectively, 
and $H$ is the hidden layer dimension size of generator. The complexity for updating discriminator is $O(K|E|d+Knd^{2})$, where $K$ is the number of layers of GCN, $|E|$ is the number of edges, and $d$ is the hidden layer dimension size of GCN. Therefore, the total time complexity of ImGAGN is $O((L-1)n_{g}H^{2}+n_{g}n_{min}^{2})+\lambda_{2}(K|E|d+Knd^{2}))$, where $\lambda_{2}$ is the number of discriminator training steps for once generator training. Furthermore, the time complexity for GCN is $O(K|E|d+Knd^2)$ and the time complexity for GraphSAGE is $O(r^K nd^2)$, where $r$ is GraphSAGE’s batch. Our method is $O((L-1)n_gH^2+n_g n_{min}^2+\lambda_{2} (K|E|d+Knd^2))$, so it means that we only have little time cost $O((L-1)n_gH^2+n_g n_{min}^2)$ which can be simplified to $O(nH^2)$ on the generator in comparison to GCN to deal with imbalanced networks because $\lambda_{2}$ is a constant number (less than 100) and it can be removed for time complexity calculation.

\section{Experiment}
In this section, we conduct the experiments on four real-world datasets to validate the effectiveness of the proposed method. Include the imbalanced node classification task, network layouts task and parameters sensitivity analysis task, aiming to answer the following research questions (RQ):
\begin{itemize}
    \item \textbf{RQ1:} How does ImGAGN perform compared with both the state-of-the-art balanced network embedding methods and imbalanced network embedding methods on imbalanced node classification task?
    \item \textbf{RQ2:} Can ImGAGN learn the node embeddings such that the representation of minority class nodes can separate from the majority class nodes?
    \item \textbf{RQ3:} How do different hyper-parameters (e.g., generated nodes ratio) influence the performance of ImGAGN?
\end{itemize}

\subsection{Experimental setup}

\subsubsection{\textbf{Datasets:}}

We conduct experiments on four publicly real-world datasets including Cora \cite{mccallum2000automating}, Citeseer \cite{giles1998citeseer}, Pubmed \cite{sen2008collective}, and DBLP \cite{tang2008arnetminer} datasets. The statistic information of the datasets is summarized in Table 1.
\begin{table}[htbp]
  \centering
  \caption{The statistic information of the network datasets.}
    \begin{tabular}{c|c|c|c|c}
    \toprule
    Datasets & Cora  & Citeseer & Pubmed & DBLP \\
    \midrule
    The number of nodes & 2708  & 3312  & 16452 & 20783 \\
    The number of edges & 5429  & 4715  & 39308 & 58188 \\
    The number of classes & 7     & 6     & 3     & 10 \\
    Feature dimension & 1433  & 3703  & 500   & 1000 \\
    Ratio of the minority class & 6.65\% & 7.52\% & 5.25\% & 1.31\% \\
    \bottomrule
    \end{tabular}%
  \label{tab:addlabel}%
\end{table}%

\begin{itemize}
    \item Cora \cite{mccallum2000automating}, Citeseer \cite{giles1998citeseer}, Pubmed \cite{sen2008collective}, and DBLP \cite{tang2008arnetminer} are the citation network datasets which consist of the nodes representing papers and the edges representing citation relationship between two papers. For each paper, a sparse bag-of-words vector is utilized as the node feature vector. For these four original datasets, the node classes (labels) are defined according to the several research topics, and each class has the roughly equal number of nodes. In our experiments, for validating the effectiveness of the proposed method on the imbalanced networks, following \cite{zhou2018sparc}, all these four balanced networks are reconstructed as the binary imbalanced networks by setting the smallest class as the minority class and the residual classes as the majority class. Specifically, taking Cora dataset for an example, there are seven classes\footnote{Neural Networks: $30.21\%$, Rule Learning: $6.65\%$, Reinforcement Learning: $8.01\%$, Probabilistic Method: $15.73\%$, Theory: $12.96\%$, Genetic Algorithm: $15.44\%$, and Case Based: $11.00\%$.} in total. Thus, the smallest class Rule Learning ($6.65\%$) is used as the minority class, and the residual classes ($93.35\%$,) are used as majority class. 
    \item For each dataset, the training, validation and testing are randomly split as ratio 7:1:2. It is worth emphasizing that the ImGAGN to generate the new balanced network is done after training/testing split, that is, the generated fake nodes would only be linked to the training minority nodes, but not the testing minority nodes.

\end{itemize}

\begin{table*}[htbp]
  \centering
  \caption{The imbalanced binary node classification results on Cora, Citeseer, Pubmed and DBLP datasets. The best results are marked in bold.}
    \begin{tabular}{c|ccc|ccc|ccc|ccc}
    \toprule
    Datasets & \multicolumn{3}{c|}{Cora} & \multicolumn{3}{c|}{Citeseer} & \multicolumn{3}{c|}{Pubmed} & \multicolumn{3}{c}{DBLP} \\
    \midrule
    Metrics & Recall & Precision & AUC   & Recall & Precision & AUC   & Recall & Precision & AUC   & Recall & Precision & AUC \\
    \midrule
    GCN   & 0.7222 & 0.8394 & 0.8973 & 0.32  & 0.5882 & 0.6388 & 0.0014 & 0.0077 & 0.8657 & 0.0363 & 0.6667 & 0.8013 \\
    GraphSAGE & 0.8056 & 0.8667 & 0.8926 & 0.32  & 0.4074 & 0.776 & 0.0025 & 0.0054 & 0.8792 & 0.0075 & 0.01  & 0.6125 \\
    GCN-SMOTE & 0.8611 & 0.6279 & 0.867 & 0.36  & 0.36  & 0.586 & 0.5376 & 0.0022 & 0.8772 & 0.5273 & 0.8947 & 0.8619 \\
    DeepWalk & 0.75  & 0.5676 & 0.883 & 0.18  & 0.012 & 0.572 & 0.3006 & 0.5327 & 0.7835 & 0.3091 & 0.0021 & 0.8648 \\
    Node2vec & 0.5833 & 0.4706 & 0.6971 & 0.107 & 0.0031 & 0.5227 & 0.3294 & 0.4474 & 0.8292 & 0.0016 & 0.0056 & 0.4893 \\
    LINE  & 0.2222 & 0.6327 & 0.8917 & 0.101 & 0.4086 & 0.8071 & 0.0982 & 0.2527 & 0.8639 & 0.0032 & 0.0734 & 0.7701 \\
    SPARC & 0.6944 & 0.8333 & 0.8822 & 0.24  & 0.6   & 0.785 & 0.0026 & 0.0039 & 0.1272 & 0.0098 & 0.0017 & 0.4525 \\
    DR-GCN & 0.7123 & 0.7899 & 0.8776 & 0.481 & 0.5614 & 0.6102 & 0.4876 & 0.2275 & 0.6714 & 0.519 & 0.7698 & 0.8122 \\
    RECT  & 0.8944 & 0.8714 & 0.8912 & 0.78  & 0.5455 & 0.7237 & 0.7624 & \textbf{0.6154} & 0.7232 & 0.8182 & \textbf{0.9016} & 0.9227 \\
    \textbf{ImGAGN} & \textbf{0.9187} & \textbf{0.893} & \textbf{0.9223} & \textbf{0.84} & \textbf{0.7121} & \textbf{0.8837} & \textbf{0.8768} & 0.5807 & \textbf{0.9086} & \textbf{0.9055} & 0.8525 & \textbf{0.9382} \\
    \bottomrule
    \end{tabular}%
  \label{tab:addlabel}%
\end{table*}%

\subsubsection{\textbf{Comparison Algorithms:}} To validate the effectiveness of the proposed method ImGAGN, we compare it with following nine state-of-the-art methods which can be grouped into two categories: balanced network embedding methods (i.e., GCN, GraphSAGE, DeepWalk, Node2vec and LINE) and imbalanced network embedding methods (i.e., GCN-SOMTE, SPARC, DR-GCN and RECT).

\begin{itemize}
    \item \textbf{GCN}: Graph convolutional network (GCN) \cite{kipf_semi-supervised_2016} is the most representative balanced network embedding method which obtains the node embedding by aggregating the neighbor nodes' features.
    \item \textbf{GraphSAGE}: GraphSAGE \cite{hamilton2017inductive} is also a representative GNN method. Unlike GCN taking the full-size neighbor nodes to obtain the node embedding, GraphSAGE adopts a fixed number of neighbor nodes for each target node to save the memory. In addition, it learns three different aggretators including Mean-aggregator, LSTM-aggregator and Pooling aggregator. We use the best performances of these three aggregator as the final results of GraphSAGE.
    \item \textbf{GCN-SMOTE}: Synthetic minority oversampling technique (SMOTE) \cite{chawla_smote_2002} is the most frequently used method to address the imbalanced classification problem by generating synthetic samples from existing minority samples. In this paper, in order to fully show the performance of the GNN methods, we incorporate the SMOTE technique into GCN for improving its performance on imbalanced network embedding problem. Specifically, we utilize the SMOTE as the data preprocessing technique only on the training set. SOMTE is used to oversample the minority class nodes to balance the classes distribution, and then the GCN is trained on the training set with balanced class distribution.
    \item \textbf{DeepWalk}: DeepWalk \cite{perozzi_deepwalk:_2014} is the most representative unsupervised network embedding method which adopts the random walk over the network to sample a set of network paths, and the neural language model (SkipGram) is applied to these network paths to obtain the node embedding.
    \item \textbf{Node2vec}: Node2vec \cite{grover_node2vec:_2016} is also an unsupervised network embedding method which obtains the node embedding by using a biased random walk strategy to preserve the homophily and structure equivalence relationships in the networks.
    \item \textbf{LINE}: LINE \cite{tang_line:_2015} obtains the network embedding by simultaneously optimizing the first-order and second-order proximities of the networks.
    \item \textbf{SPARC}: SPARC \cite{zhou_sparc:_2018} is an imbalanced network embedding method. It obtains the imbalanced embedding in a mutually way, which can jointly predict the minority class and the neighbor context in the networks.
    \item \textbf{DR-GCN}: DR-GCN \cite{shi_multi-class_2020} is also a GCN based imbalanced network embedding method, and it proposes to use conditioned adversarial training to enhance the separation of different classes. In addition, the distribution alignment training is applied to balance between the majority nodes and the minority nodes.
    \item \textbf{RECT}: RECT \cite{wang_network_2020} is the state-of-the-art imbalanced network embedding method which is a variant of GNN. It obtains the imbalanced network embedding by learning the knowledge of class-semantic information in the networks.
\end{itemize}

\begin{figure*}[t]
    \centerline{\includegraphics[width=0.85\textwidth]{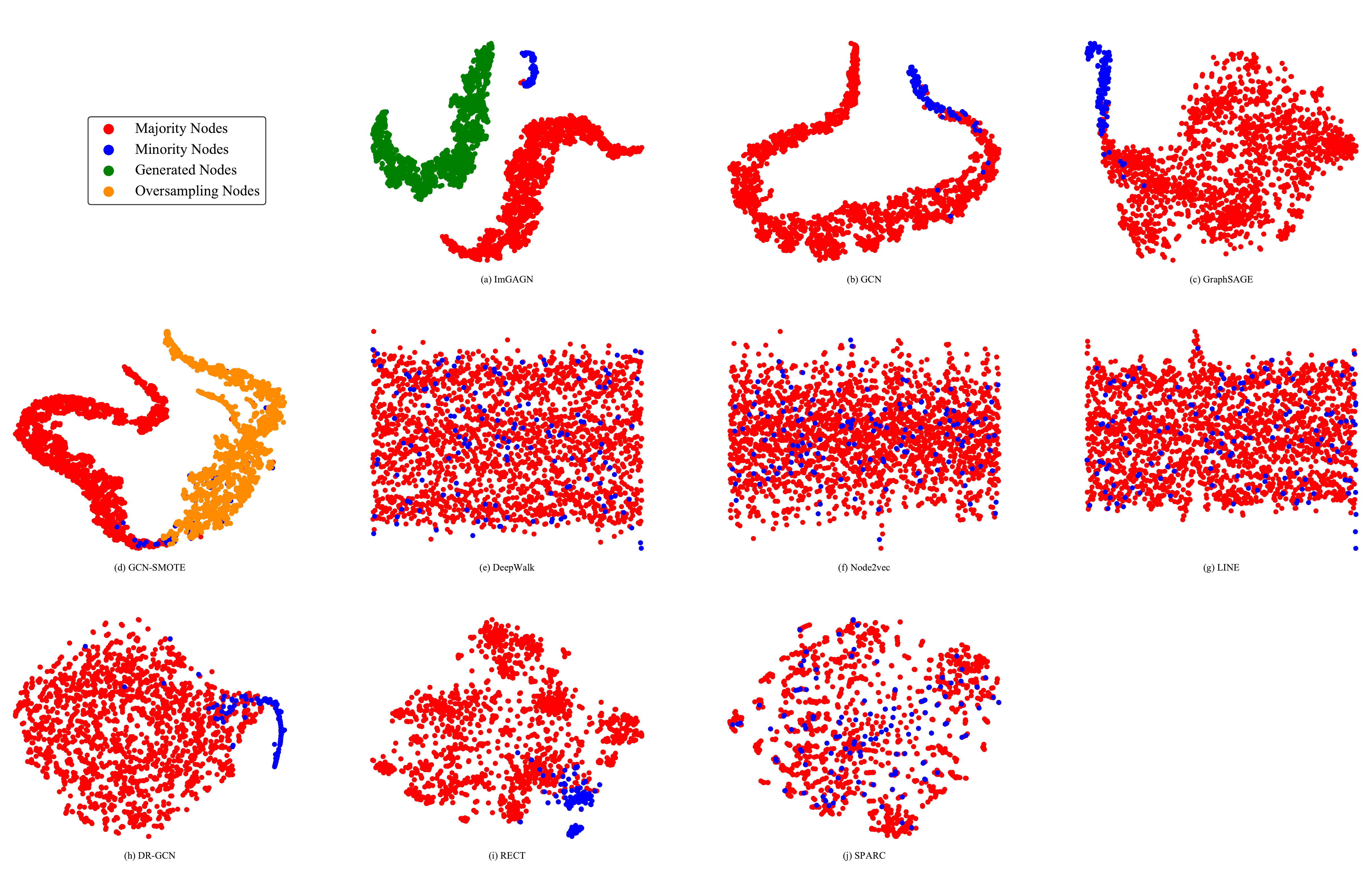}}
    \caption{The 2-dimensional imbalanced network layout with t-SNE on Cora dataset. The red circles represent the majority nodes of the original networks. The blue circles represent the minority nodes of the original networks. The yellow circles represent the minority nodes generated by SMOTE. The green circles represent the minority nodes generated by the proposed ImGAGN. The proposed ImGAGN is capable of discriminating between the real nodes (i.e., the blue circles and red circles) and the generated fake nodes (i.e., the green circles), and between the minority nodes (i.e., the blue nodes) and the majority nodes (i.e., the red nodes).}
    \label{fig}
\end{figure*}

\subsubsection{\textbf{Parameters:}} All the codes we used are provided by authors. For GCN, following \cite{kipf_semi-supervised_2016}, the number of layers of the networks is set $K=2$. For GraphSAGE, we set $K=2, S_{1}=5, S_{2}=5$ according to the author suggesting. For GCN-SMOTE, the number of generated minority samples by SMOTE is equal to the difference between the majority and minority nodes of the training set. For DeepWalk, we adopt the default hyperparameters (i.e., window size $win=10$, walk length $len = 40$ and the number of walks $t=90$). For Node2vec, we optimize its hyperparameters by a grid search over $p,q \in \{0.25,0.50,1,2,4\}$. For LINE, the hyperparameter negative samples $ns=5$. For SPARC, the length of random walk sequences $\mu =10$. Moreover, the embedding dimension of unsupervised network embedding methods (i.e., DeepWalk, Node2vec and LINE) are set as $d=128$, and the logistic regression classifier is employed to evaluate the node embedding. For semi-supervised network embedding methods (i.e., GCN, GCN-SMOTE, GraphSAGE, SPARC and RECT), we use the outputs of their last hidden layer as the node embedding (the embedding dimension is also $128$).

The hyperparameters of our proposed method ImGAGN are set as follows. For generator, it consists of 3 fully connected layers with $100$ units in input layer and $200$ units in hidden layer. The number of units of output layer is equal to the difference between the majority class and minority class of the training set. $Tanh()$ is utilized as the activation function. For discriminator, it consists of the two-layer GCN followed by a softmax function, and ReLU \cite{glorot_deep_nodate} is utilized as the activation function. In addition, we perform generator and discriminator updates in $1:100$ ratio, and Adam SGD optimizer \cite{kingma_adam_2017} is utilized as the optimizer throughout the experiments.

\subsubsection{\textbf{Repeatability:}}
All the methods are run on a single machine with 14 CPU cores at 2.60GHZ and 2 Tesla P100 GPU with 32G memory using 1 thread. 

\subsection{Imbalanced binary node classification (RQ1)}
To answer the RQ1 (i.e., how does ImGAGN perform compared with both the state-of-the-art balanced and imbalanced network embedding), we first conduct imbalanced binary node classification experiment on the four real-world network datasets. Three common classification metrics are used to evaluate the performance for all algorithms. Include: (1) recall, which measures the ratio of correctly classified nodes of all minority test nodes. (2) precision, which measures the ratio of correctly classified nodes of all predicted minority nodes (3) AUC scores, which measures model performance at all classification thresholds. We run experiments 10 times and use average scores for each metric. The experimental results are shown in Table 2.

From experimental results, in general, we can observe that: 
\begin{itemize}
    \item The proposed method ImGAGN substantially outperforms all comparison methods with respect to recall and AUC on all datasets and is comparable with RECT in terms of precision on Pubmed and DBLP datasets, nonetheless, ImGAGN improves significantly by 11.44\% and 8.73\% on these two datasets respectively in terms of recall which is usually more important than precision in many minority class classification problems, such as rare disease prediction \cite{schubach2017imbalance}. Thus, the overall performance could validate the effectiveness of the proposed method.
    \item As expected, the imbalanced network embedding methods (i.e., GCN-SOMTE, SPARC, RECT, DR-GCN and ImGAGN) achieve better performance than the balanced network embedding methods (i.e., GCN, GraphSAGE, DeepWalk, Node2vec and LINE) in most case. It is reasonable since the former methods focus more on label learning of the minority class samples.
    \item The GCN-SMOTE achieves better performance than original GCN, which shows that simple oversampling technique is capable of improving original GCN performance on imbalanced network data. The proposed method ImGAGN can also be thought of as an oversampling technique due to the operations of GraphGenerator. However, it obtains better performance than GCN-SMOTE, the improvements could be attributed to the GraphGenerator could well capture both implicit topological structure distribution and nodes' attribute distribution of the minority nodes.
\end{itemize}

\subsection{Network layout (RQ2)}
To answer the RQ2 (i.e., can ImGAGN learn the node embeddings such that the representation of minority class nodes can separate from the majority class nodes? ), we visualize the network layout in the embedding space, and we take Cora dataset for an example. Specifically, we firstly learn the nodes embedding in a 128-dimensional vector space for different network embedding methods, and then employ the t-SNE \cite{maaten2008visualizing} to map the 128-dimensional into the 2-dimensional space for visualization. The experimental results are shown in Figure 3. From the experimental results, in general, we can observe that:
\begin{itemize}
    \item The proposed ImGAGN is well capable of discriminating between the real nodes (i.e., the blue circles and red circles) and the generated fake nodes (i.e., the green circles), and also between the minority nodes (i.e., the blue nodes) and the majority nodes(i.e., the red nodes), which validates ImGAGN is able to capture the latent representation of the minority nodes and majority nodes. We attribute such performance to the architecture of the GAN-based methods, that is, the loss functions of GraphGenerator (i.e., equation (4)) and discriminator (i.e., equation (6)) explicitly learn the discrimination between the real nodes and fake nodes, and also between minority nodes and majority nodes. 
    \item Generally speaking, the end-to-end semi-supervised network embedding methods (i.e., GCN, GCN-SMOTE, GraphSAGE, SPARC, RECT and ImGAGN) can better discriminate the majority and the minority classes than the unsupervised network embedding methods (i.e., DeepWalk, Node2vec and LINE). One explanation is that semi-supervised methods leverage both the nodes' features and label information to learn separable representation of the minority and majority nodes. 
\end{itemize}

\subsection{Parameters sensitivity analysis (RQ3)}
To answer the RQ3 (i.e., How do different hyper-parameters influence the performance of ImGAGN?). We conduct the imbalanced node classification experiments on Cora and DBLP datasets and report the performance changes with respect to two crucial hyperparameters of the ImGAGN. One is $\lambda_{1}$ which is the ratio of the number of all training minority nodes (i.e., the original minority nodes and the generated minority nodes in training set) to the number of majority nodes, and we vary it from $0$ to $1$ with step size $0.1$. Another is $\lambda_{2}$ which is the number of discriminator training steps to for once generator training), and we vary it from $10$ to $100$ with step size $10$. The experimental results are shown in Figure 4 and Figure 5. From experimental results, in general, we can observe that:
\begin{itemize}
    \item The imbalanced node classification performance, especially for recall, increases with the increase of training minority nodes ratio $\lambda_{1}$ and then tend to flat. One explanation is that when $\lambda_{1}$ is small, the training network is still classes imbalanced, which leads to bad classification performance. It is worth nothing that $\lambda_{1}=0$ degenerates our method to GCN, thus the performance with $\lambda_{1}=0$ also demonstrates that the proposed GraphGenerator could improve the GCN performance on imbalanced networks. 
    \item The performance increases with the increase of discriminator training steps $\lambda_{2}$ and then tend to flat. It is reasonable since the discriminator needs a certain number of training steps to learn the node embeddings.
    \item Particular speaking, we found the proposed method ImGAGN could achieve high performance with $\lambda_{1} > 0.7$ and $\lambda_{2}>50$.
\end{itemize}

\begin{figure}[t]
\centering
\includegraphics[width=0.48\textwidth]{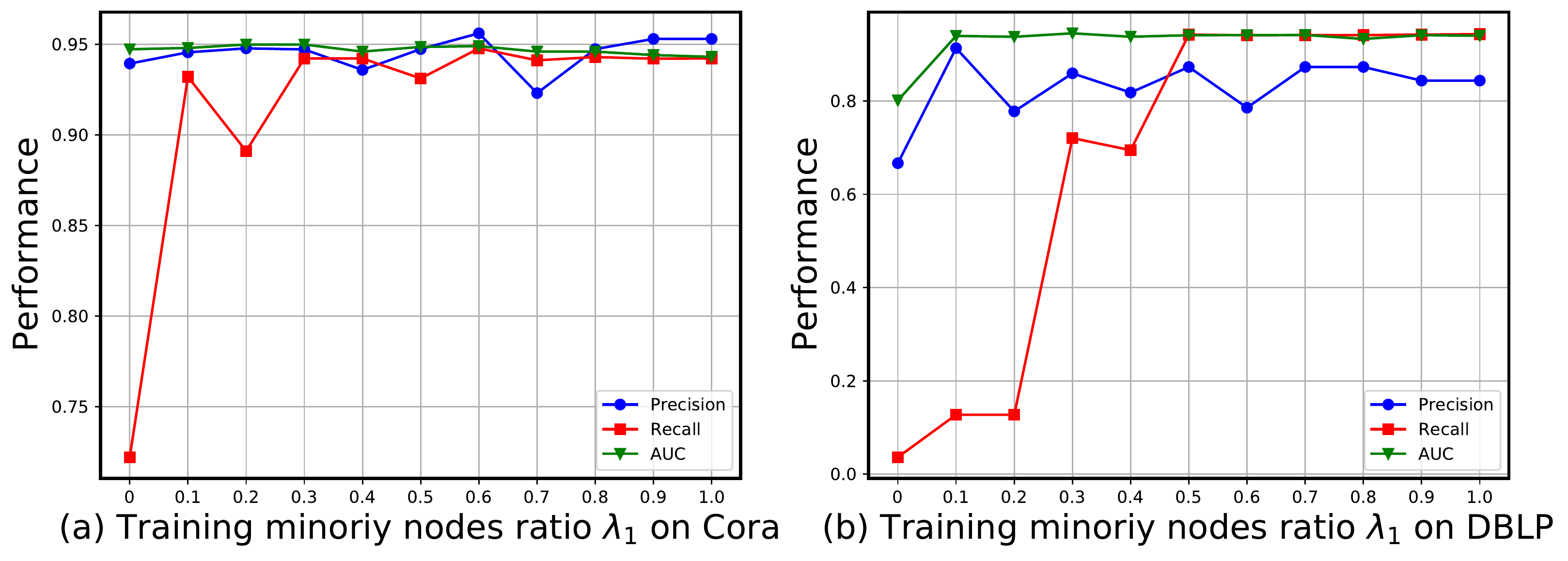} % Reduce the figure size so that it is slightly narrower than the column.
\caption{Hyper-parameter sensitivity analysis of the training minority nodes ratio $\lambda_{1}$.}
\label{fig2}
\end{figure}

\begin{figure}[t]
\centering
\includegraphics[width=0.48\textwidth]{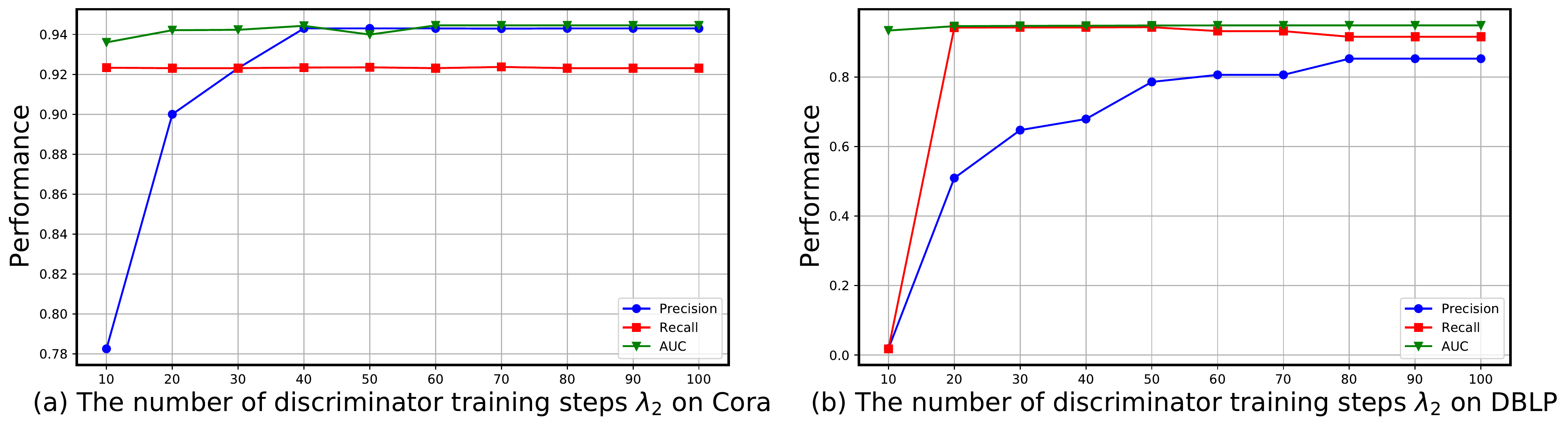} % Reduce the figure size so that it is slightly narrower than the column.
\caption{Hyper-parameter sensitivity analysis of the discriminator training steps $\lambda_{2}$.}
\label{fig2}
\end{figure}

\section{Conclusion}
In this paper, to address the imbalanced network embedding problem, we proposed a semi-supervised network embedding method ImGAGN, which utilized a GraphGenerator to simulate both the minority class nodes’ attribute distribution and network topological structure distribution. It generated a set of synthetic minority nodes such that the number of nodes in different classes can be balanced. Then GCN discriminator was trained to discriminate between real nodes and fake nodes, and also between minority nodes and majority nodes. The extensive comparative studies, including the imbalanced node classification, network layouts and hyper-parameters sensitivity analysis, are conducted to validate the effectiveness of the proposed method. The empirical evaluation on four real-world datasets demonstrated that the proposed ImGAGN could outperform the state-of-the-art imbalanced network embedding algorithms on imbalanced node classification task in most cases with respect to recall, precision and AUC. In addition, the visualization results showed that the ImGAGN is capable of learning the node embedding such that the minority class nodes can separate from the majority class nodes.
% and we also intend to apply ImGAGN to the networks with more properties, such as heterogeneous information network, and dynamic network.

\section{Acknowledgement}
This work is partially supported by the Science and Technology Innovation Committee Foundation of Shenzhen under the Grant No. JCYJ20200109141235597 and ZDSYS201703031748284, National Science Foundation of China under grant number 61761136008, Shenzhen Peacock Plan under Grant No. KQTD2016112514355531, Program for Guangdong Introducing Innovative and Entrepreneurial Teams under grant number 2017ZT07X386, ARC Discovery Project under the Grant No. DP190101985 and ARC Training Centre for Information Resilience under the Grant No. IC200100022.

\bibliographystyle{ACM-Reference-Format}
\bibliography{main.bib}

\end{document}